\documentclass{article}
\usepackage{ijcai26}

\usepackage{times}
\usepackage{soul}
\usepackage{url}
\usepackage[hidelinks]{hyperref}
\usepackage[utf8]{inputenc}
\usepackage[small]{caption}
\usepackage{graphicx}
\usepackage{amsmath}
\usepackage{amsthm}
\usepackage{booktabs}

\newcommand{\bestnum}[1]{$\pmb{#1}$}
\newcommand{\secondnum}[1]{\underline{#1}}

\ifdefined
\fi

\title{Low-Cost Labels, Reliable Choices:\\
Rollout-Calibrated Hyper-Heuristics for Job Shop Scheduling}

\author{
Junhao Wei$^{1,2}$
\and Yanxiao Li$^{1}$
\and Yifu Zhao$^{1}$
\and Zhenhong Peng$^{3}$
\and Baili Lu$^{1}$
\and Dexing Yao$^{1}$
\and Haochen Li$^{1}$
\and Qinbin He$^{1}$
\and Sio-Kei Im$^{4}$
\and Yapeng Wang$^{1}$\And
Xu Yang$^{1}$\thanks{Corresponding author: xuyang@mpu.edu.mo}\\
\affiliations
$^{1}$Faculty of Applied Sciences, Macao Polytechnic University, Macao, 999078, China\\
$^{2}$Pazhou Lab (Huangpu), Guangzhou, 510555, China\\
$^{3}$College of Animal Science and Technology, Zhongkai University of Agriculture and Engineering, Guangzhou, 510225, China\\
$^{4}$Macao Polytechnic University, Macao, 999078, China\\
}

\begin{document}

\maketitle

\begin{abstract}
Learning-assisted hyper-heuristics can select among dispatching rules while
preserving the feasibility and interpretability of constructive Job Shop
Scheduling Problem (JSSP) heuristics. Their main computational cost lies in
label generation rather than model fitting, since each supervised label usually
requires rolling out candidate rules from a partial schedule. We study this
label-cost problem together with a reliability problem: a learned selector
should not switch away from a strong default rule unless the predicted gain is
credible. The proposed selector uses regret-normalized rollout labels, a
contextual KNN uncertainty estimate, and a gate that acts only when the
predicted improvement exceeds an uncertainty-adjusted margin. We also vary
rollout depth and breadth to measure the cost-quality trade-off. On synthetic
JSSP instances, the gated selector achieves the lowest mean RPD among learned
selectors, remains close to the best fixed dispatching rule, and reduces
Random-HH mean RPD by more than an order of magnitude.
\end{abstract}

\section{Introduction}
Selection hyper-heuristics choose among low-level rules while a schedule is
being built. For JSSP, this gives a useful middle ground between fixed
dispatching rules and end-to-end learned constructors: the action space remains
small, feasibility is maintained by construction, and the rule choices are
directly interpretable.

This setting falls within the LEAD themes of hyper-heuristics and
operator/algorithm selection. The learner does not design a new dispatching
rule; it decides which existing rule should be trusted at a given partial
schedule. This separation isolates rule selection from the separate problem of
inventing new search operators.

The main bottleneck is supervision. A common labeler rolls out every candidate
rule from a partial schedule and records the terminal makespan. This gives
direct supervision, but a full rollout costs
$O(|\mathcal{H}|T)$ simulations for a state with $T$ remaining decisions. A
second issue appears at inference: a learned score can be slightly better than
the default rule for accidental reasons, and always following the lowest score
can lose to a strong fixed rule.

Our formulation ties these issues together by storing rollout outcomes as
per-state regret rather than raw or lower-bound-normalized makespan, so the
target reflects the local ranking of rules. The selector then uses a variance
estimate from a contextual KNN regressor to switch away from the default rule
only when the predicted margin is larger than the uncertainty penalty. To make
the supervision cost explicit, we vary rollout depth and breadth and report the
resulting cost-quality trade-off.

The scope differs from specialized JSSP solvers, deep-RL dispatchers such as
L2D~\cite{zhang2020l2d}, and program search systems such as
FunSearch~\cite{romera-paredes2024funsearch} and ReEvo~\cite{ye2024reevo}.
The contribution is a reproducible CPU-only selector and evaluation protocol
for studying label cost and conservative switching in learning-assisted
hyper-heuristics.

The experiments therefore measure the value of regret labels, the reliability
benefit of uncertainty gating, and the amount of rollout depth or breadth that
can be removed before the selector loses its advantage. This evaluation matches
a practical deployment choice about how much simulation budget should be spent
before a learned selector is used.

\section{Related Work}

Selection hyper-heuristics are the closest line of work to our setting because
they choose low-level heuristics for combinatorial optimization problems
\cite{burke2013hyper}. Online-learning variants use multi-armed bandits
\cite{fialho2010analyzing} or reinforcement learning
\cite{ozcan2010reinforcement} to control rule choice at each decision point.
Prior work on dispatching rule selection for JSSP includes the
imitation-learning selector of Ingimundardottir and Runarsson~\cite{ingimundardottir2018discovering}
and RANFORS~\cite{jun2019learning}, where rule priorities are learned from
instance features.

Learning-based scheduling provides a parallel line of work. Recent studies
treat JSSP as a sequential decision process, as in L2D~\cite{zhang2020l2d}
and ScheduleNet~\cite{park2021schedulenet}, often trained with policy
gradients. Related neural combinatorial-optimization methods such as
POMO~\cite{kwon2020pomo} also illustrate the dependence of learned policies on
training distributions. In contrast, our selector uses rollout labels,
engineered state features, and a CPU-only KNN regressor.

Rollout policies supply the label-generation mechanism used here.
Bertsekas-style rollout policies \cite{bertsekas2013rollout} estimate the
value of an action by simulating a base policy. We borrow this idea for label
generation rather than online control, which lets us amortize the simulation
cost across many decision points. Surrogate-based hyper-heuristics
\cite{horn2016multi,branke2016model} similarly trade simulation cost for
generalization. Our Pareto sweep makes the rollout cost visible instead of
treating it as a fixed preprocessing step.

LLM- and agentic-driven algorithm design is related but addresses a different
part of the pipeline. LLM-based systems such as EoH~\cite{liu2024eoh},
FunSearch~\cite{romera-paredes2024funsearch}, and ReEvo~\cite{ye2024reevo}
can generate or refine heuristic code directly. Our method keeps the rule pool
fixed and studies when a compact selector should trust its predictions.

The work is also connected to metaheuristic design and applications, where
LEAD studies often improve search algorithms by designing new movement
patterns, update equations, or hybrid prediction pipelines. Recent examples
include TSWOA~\cite{wei2025tswoa} and geometric WOA~\cite{wei2026geometric}
for engineering design, MRBMO~\cite{lu2025mrbmo} for numerical optimization,
and metaheuristic-assisted prediction models such as
NAWOA-XGBoost~\cite{wei2026nawoa} and
ASKSSA-CNN-BiLSTM~\cite{li2026askssa}. These studies modify the optimizer
itself. Our work is complementary: the low-level rules are fixed, and learning
is used only to select among them.

\section{Method}\label{sec:method}

\subsection{JSSP and Selection Hyper-Heuristics}
A JSSP instance is a tuple $(\mathcal{J}, \mathcal{M}, \pi, p)$ where each job
$j \in \mathcal{J}$ has an ordered operation sequence $\pi_j$ over machines
$\mathcal{M}$ with processing times $p$. A serial schedule generator
\cite{giffler1960algorithms} builds a feasible schedule by repeatedly selecting
the next operation from the set of ready jobs. The makespan
$C_{\max}$ is the largest job completion time. We use the classical priority
dispatching rules covered in standard scheduling-rule surveys and job-shop
benchmarks~\cite{panwalkar1977survey,holthaus1997efficient}. The rule pool
$\mathcal{H}$ contains \textsc{SPT}, \textsc{LPT}, \textsc{MWKR},
\textsc{LWKR}, \textsc{MOPNR}, \textsc{FIFO}, and \textsc{Random}. These rules
choose by shortest or longest processing time, most or least remaining work,
most remaining operations, first-in-first-out order, or a uniform random
choice, respectively. A selection hyper-heuristic chooses one
$h \in \mathcal{H}$ at every decision point as a function of the current state
$s$.

\subsection{Regret-Normalized Rollout Labels}
\label{sec:method:regret}
Given a partial state $s$, we evaluate each candidate $h$ by completing the
schedule from $s$ using $h$ for the next step and then continuing the rollout.
Let $m(s, h)$ denote the resulting makespan. The classical normalized label is
$m(s,h) / \mathrm{LB}(I)$ where $\mathrm{LB}$ is the instance lower bound. We
instead use the per-state regret
\begin{equation}
r(s, h) = \frac{m(s, h) - \min_{h' \in \mathcal{H}_s} m(s, h')}
                 {\min_{h' \in \mathcal{H}_s} m(s, h')},
\label{eq:regret}
\end{equation}
where $\mathcal{H}_s \subseteq \mathcal{H}$ is the candidate subset evaluated
at $s$. By construction $r(s, h) \ge 0$ with $r(s, h) = 0$ for the best rule.
The target is therefore a local ranking signal instead of an absolute makespan
scale.

\subsection{Uncertainty-Gated Selection}
\label{sec:method:gated}
We fit a contextual KNN regressor on $(s, h)$ pairs whose features are
described in Section~\ref{sec:method:features}. For test state $s$ we compute,
for each $h \in \mathcal{H}$:
\begin{align}
\hat{r}(s, h) &= \frac{\sum_{i \in N_k(s,h)} w_i \, y_i}{\sum_i w_i}, &
\hat{\sigma}(s, h) = \mathrm{std}_{i \in N_k(s,h)}(y_i),
\end{align}
where $N_k$ is the $k$-nearest training neighbor set and $w_i = 1/(d_i + \epsilon)$.
Let $h^{\star} = \arg\min_h \hat{r}(s, h)$ and $h_0$ be a fixed default rule
(in our experiments, the best fixed rule on the training set).
The \emph{gated} policy applies $h^{\star}$ only when the expected
improvement over $h_0$ exceeds a confidence margin:
\begin{equation}
\pi_{\mathrm{gated}}(s) =
\begin{cases}
h^{\star} & \text{if } \hat{r}(s, h_0) - \hat{r}(s, h^{\star}) > \lambda \, \hat{\sigma}(s, h^{\star}), \\
h_0 & \text{otherwise.}
\end{cases}
\label{eq:gate}
\end{equation}
The hyperparameter $\lambda$ trades aggressiveness for reliability: $\lambda = 0$
recovers $\arg\min_h \hat{r}$, while large $\lambda$ collapses to the default
rule. We also report an LCB-style variant
$h_{\mathrm{lcb}} = \arg\min_h \hat{r}(s, h) - \lambda \hat{\sigma}(s, h)$.

\subsection{Rollout-Budget Trade-off}
\label{sec:method:pareto}
The cost of label generation is governed by the rollout breadth
$b = |\mathcal{H}_s|$, which determines how many candidate rules are evaluated
per state, and the rollout depth $\kappa$, which determines how long the
candidate rule is followed before completion resumes with the default rule
$h_0$. The full-rollout setting with $\kappa = \infty$ and
$b = |\mathcal{H}|$ is the most expensive case. By varying $(\kappa, b)$ on a
small grid, we trace a Pareto curve relating label generation cost, measured by
rollout steps or wall-clock seconds, to the test RPD of the learned selector.

\subsection{State Features}\label{sec:method:features}
The state representation contains 35 instance- and state-level features,
covering instance size, processing-time statistics, machine load imbalance,
schedule progress, ready-set processing times, remaining work, and machine and
job readiness moments, together with a 7-dimensional one-hot encoding of the
candidate heuristic. All features are $z$-normalized at training time and the
same normalization is reused at inference.

\subsection{Training Pipeline}
For each training instance, we sample partial schedules by following uniformly
random dispatching rules. At each sampled state, the labeler evaluates a subset
of candidate rules, applies each candidate for the next decision, and completes
the remaining schedule according to the rollout depth setting. Full depth means
that the candidate rule is followed to terminal completion; finite depth means
that the rollout returns to the default rule after $\kappa$ candidate-guided
steps. The recorded makespans are converted to either normalized targets or the
regret target in Eq.~\eqref{eq:regret}.

The default rule $h_0$ is chosen on the training split as the fixed rule with
the lowest mean makespan. This makes the gating test conservative: the learned
selector must justify switching away from a rule that is already competitive on
the deployment distribution. During inference, the KNN model scores every
candidate rule at each decision point and the gate in Eq.~\eqref{eq:gate}
decides whether to switch away from $h_0$.

\section{Experiments}\label{sec:experiments}

\subsection{Setup}
\label{sec:exp:setup}
We evaluate on synthetic JSSP instances at three scales: $6 \times 6$,
$10 \times 10$, and $15 \times 10$ (jobs $\times$ machines). Each instance
draws an independent random machine permutation per job and processing times
uniformly from $[1, 99]$. We use 150 training instances and 40 held-out test
instances per scale.

The selector is a CPU-based contextual KNN regressor with $k = 7$. Its fitting
step normalizes the rollout feature vectors and stores their targets; nearest
neighbor distances are computed with NumPy on CPU during inference. The rollout
labels are collected from 25 sampled states per training instance and three
exploration trajectories per instance under a uniform-random rule mixture. The
default training setting evaluates all seven candidate rules at each sampled
state and rolls each candidate to terminal completion, with breadth $b = 7$ and
depth $\kappa = \infty$, unless Section~\ref{sec:exp:pareto} explicitly varies
the rollout budget. The gating default $h_0$ is selected automatically as the
best fixed rule on the training set, which is typically \textsc{FIFO} or
\textsc{MOPNR} at our scales.

The learned configurations are \emph{Norm-Argmin}, which uses
lower-bound-normalized rollout labels with $\arg\min$ selection,
\emph{Regret-Argmin}, and \emph{Regret-Gated} at $\lambda = 1$. We compare them
with the same dispatching-rule baselines \textsc{SPT}, \textsc{LPT},
\textsc{MWKR}, \textsc{LWKR}, \textsc{MOPNR}, and \textsc{FIFO}
\cite{panwalkar1977survey,holthaus1997efficient}; a \emph{Random-HH} baseline
that chooses a rule uniformly at random at each decision and is averaged over
five seeds; and the \emph{Oracle-Fixed} reference, which selects the best fixed
rule \emph{per instance, after the fact}. We report mean relative percentage
deviation (RPD) against this hindsight reference, the median RPD, and the
number of test instances on which a method is best. Oracle-Fixed is used only
to normalize performance and is not a deployable scheduling policy.
Experiments were run on Ubuntu 20.04 with an Intel Xeon E5-2698 v4 CPU
(40 logical cores), 251 GiB RAM, and Python 3.13; no GPU was used for label
generation, KNN fitting, or inference.

\subsection{Example Schedule Visualization}
\label{sec:exp:gantt}
\begin{figure}[t]
\centering
\includegraphics[width=\linewidth]{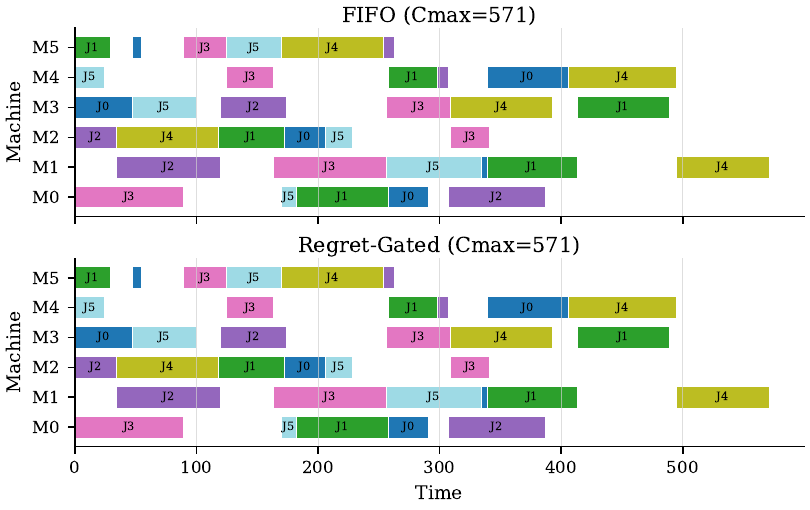}
\caption{Example 6\(\times\)6 schedules produced by FIFO and Regret-Gated.}
\label{fig:gantt}
\end{figure}
Figure~\ref{fig:gantt} gives a schedule-level view of the policies.
Each bar is one operation, the row is the assigned machine, and the color
marks the job. The visualization is not used as an optimization metric; it
checks that the learned selector produces a feasible JSSP schedule.

\subsection{Main Results}
\label{sec:exp:main}
\begin{table*}[t]
\centering
\caption{Main test results. Oracle-Fixed is a hindsight reference used to compute RPD, not a deployable method. Arrows indicate metric direction; bold and underlined values mark the best and second-best non-oracle entries.}
\label{tab:main}
\small
\begin{tabular}{lcccccc}
\toprule
 & \multicolumn{2}{c}{6x6} & \multicolumn{2}{c}{10x10} & \multicolumn{2}{c}{15x10} \\
\cmidrule(lr){2-3}\cmidrule(lr){4-5}\cmidrule(lr){6-7}
Method / reference & mean RPD$\downarrow$ & wins$\uparrow$ & mean RPD$\downarrow$ & wins$\uparrow$ & mean RPD$\downarrow$ & wins$\uparrow$ \\
\midrule
Oracle-Fixed (ref.) & 0.00 & 33 & 0.00 & 32 & 0.00 & 30 \\
\midrule
\textbf{Regret-Gated (ours)} & \secondnum{2.85} & \secondnum{19} & \secondnum{1.83} & \secondnum{23} & \secondnum{0.84} & \secondnum{25} \\
Regret-Argmin & 4.94 & 16 & 3.15 & 21 & 1.07 & \secondnum{25} \\
Norm-Argmin & 5.30 & 17 & 3.24 & 20 & 1.10 & 24 \\
FIFO & \bestnum{2.15} & \bestnum{22} & \bestnum{1.16} & \bestnum{27} & \bestnum{0.78} & \bestnum{27} \\
MOPNR & 7.77 & 11 & 9.99 & 7 & 9.04 & 3 \\
MWKR & 18.33 & 1 & 22.59 & 0 & 22.65 & 0 \\
Random-HH & 36.68 & 0 & 47.42 & 0 & 54.54 & 0 \\
SPT & 103.13 & 0 & 197.56 & 0 & 233.02 & 0 \\
LPT & 111.80 & 0 & 224.92 & 0 & 263.16 & 0 \\
LWKR & 165.37 & 0 & 310.26 & 0 & 368.64 & 0 \\
\bottomrule
\end{tabular}
\end{table*}

\begin{figure}[t]
\centering
\includegraphics[width=\linewidth]{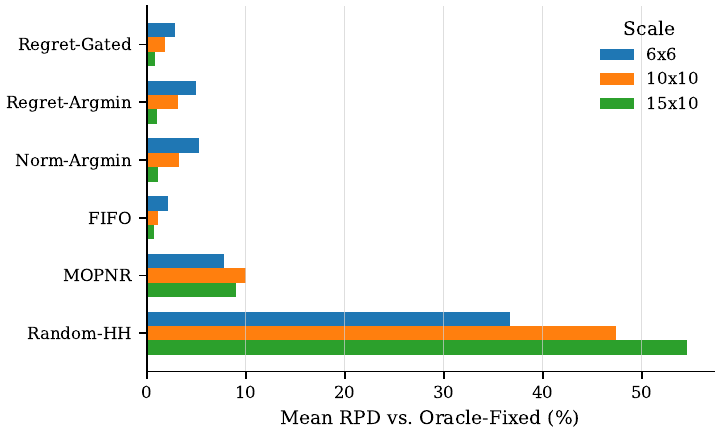}
\caption{Mean RPD for learned selectors and key baselines. Lower is better.}
\label{fig:main}
\end{figure}
Table~\ref{tab:main} and Figure~\ref{fig:main} summarize our main results; the
figure focuses on the learned selectors and key baselines.
\emph{Regret-Gated} has the lowest mean RPD among learned selectors at all
three scales. It remains close to the best fixed rule in the table,
\textsc{FIFO}, with gaps of 0.70, 0.67, and 0.06 mean RPD on $6\times6$,
$10\times10$, and $15\times10$, respectively. It also improves over the two
ungated selectors and reduces Random-HH mean RPD from 36.68--54.54 to
0.84--2.85.

\subsection{Ablation on Label Targets and Selection Policies}
\label{sec:exp:ablation}
\begin{table}[t]
\centering
\caption{Ablation over label targets and selection policies on 10x10. Arrows indicate metric direction; bold marks the best mean RPD.}
\label{tab:ablation}
\scriptsize
\begin{tabular}{@{}lccc@{}}
\toprule
Configuration & mean RPD$\downarrow$ & median RPD$\downarrow$ & wins$\uparrow$ \\
\midrule
\textbf{regret-gated-l1.0} & \bestnum{1.13} & 0.00 & 12 \\
normalized-gated-l1.0 & 1.16 & 0.00 & 12 \\
normalized-gated-l2.0 & 1.16 & 0.00 & 12 \\
regret-gated-l2.0 & 1.16 & 0.00 & 12 \\
normalized-gated-l0.5 & 1.39 & 0.00 & 12 \\
regret-gated-l0.5 & 1.72 & 0.00 & 12 \\
normalized-lcb-l1.0 & 1.75 & 0.00 & 11 \\
normalized-argmin & 2.05 & 0.00 & 11 \\
regret-argmin & 2.12 & 0.00 & 12 \\
normalized-lcb-l0.5 & 2.38 & 0.00 & 11 \\
regret-lcb-l0.5 & 3.34 & 1.65 & 9 \\
regret-lcb-l1.0 & 8.22 & 7.99 & 2 \\
normalized-lcb-l2.0 & 11.71 & 12.97 & 1 \\
regret-lcb-l2.0 & 40.54 & 40.07 & 0 \\
\bottomrule
\end{tabular}
\end{table}

\begin{figure}[t]
\centering
\includegraphics[width=\linewidth]{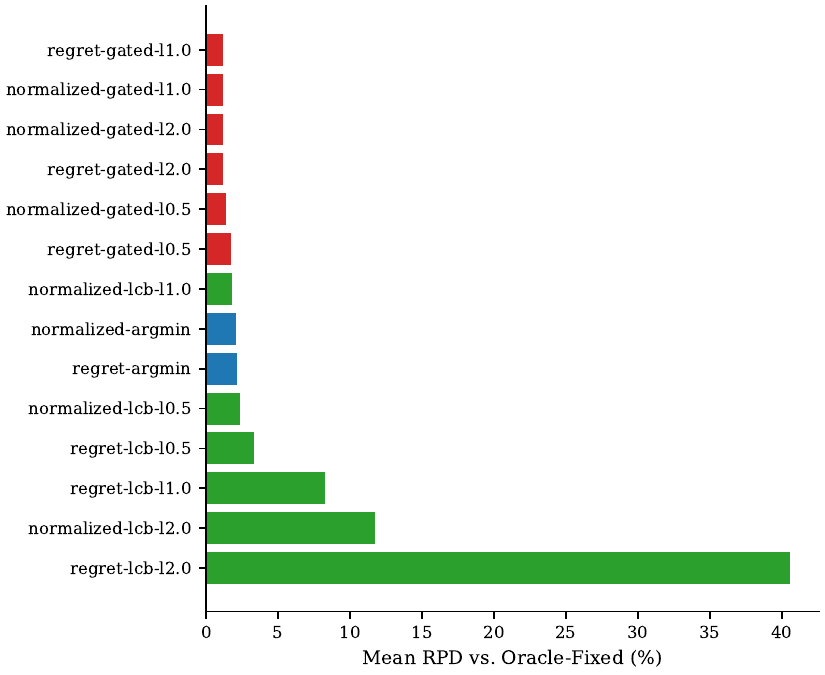}
\caption{Ablation over label targets, selection policies, and confidence thresholds.}
\label{fig:ablation}
\end{figure}
The ablation in Figure~\ref{fig:ablation} varies the label target, selection
policy, and confidence parameter across normalized and regret labels, argmin,
LCB, and gated selection, and $\lambda \in \{0.5, 1.0, 2.0\}$. The lowest mean
RPD is obtained by regret-gated selection with $\lambda=1.0$. Gating is more
stable than always switching: the argmin variants are around 2.0 mean RPD,
whereas the top gated variants are near 1.1. LCB performs worse in this setup,
suggesting that uncertainty is more useful as a switching test than as a direct
score bonus.

\subsection{Rollout-Budget Trade-off}
\label{sec:exp:pareto}
\begin{table}[t]
\centering
\caption{Rollout label cost and test quality on 10x10. Arrows indicate metric direction; bold marks the best RPD.}
\label{tab:pareto}
\scriptsize
\begin{tabular}{@{}cccccc@{}}
\toprule
depth & b & time (s)$\downarrow$ & rollouts$\downarrow$ & steps$\downarrow$ & RPD$\downarrow$ \\
\midrule
\textbf{full} & \textbf{full} & 5.90 & 8400 & 433377 & \bestnum{1.34} \\
full & 3 & 2.46 & 3600 & 185733 & 3.33 \\
full & 5 & 4.08 & 6000 & 309555 & 2.40 \\
1 & full & 5.83 & 8400 & 16716 & 25.66 \\
1 & 3 & 3.07 & 3600 & 7164 & 17.35 \\
1 & 5 & 3.87 & 6000 & 11940 & 21.84 \\
3 & full & 4.87 & 8400 & 33089 & 12.56 \\
3 & 3 & 3.17 & 3600 & 14181 & 10.70 \\
3 & 5 & 4.94 & 6000 & 23635 & 13.15 \\
5 & full & 4.85 & 8400 & 49161 & 10.53 \\
5 & 3 & 3.60 & 3600 & 21069 & 10.47 \\
5 & 5 & 3.49 & 6000 & 35115 & 10.12 \\
10 & full & 5.24 & 8400 & 88333 & 9.51 \\
10 & 3 & 2.23 & 3600 & 37857 & 12.43 \\
10 & 5 & 4.05 & 6000 & 63095 & 9.14 \\
\bottomrule
\end{tabular}
\end{table}

\begin{figure}[t]
\centering
\includegraphics[width=\linewidth]{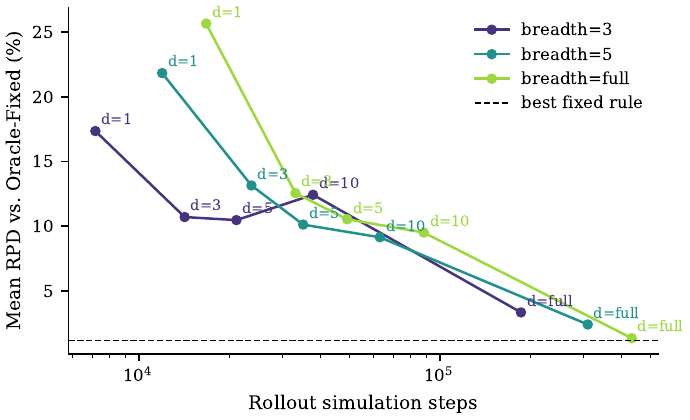}
\caption{Rollout cost vs.\ test RPD on 10\,\(\times\)\,10 JSSP.}
\label{fig:pareto}
\end{figure}
Figure~\ref{fig:pareto} traces the cost-quality frontier of rollout-based
label generation. Full rollouts achieve the lowest RPD, whereas very shallow
rollouts with $\kappa=1$ yield substantially higher RPD even when all rules are
evaluated. Intermediate depths reduce simulation steps substantially but leave
a measurable quality gap, and breadth reduction becomes useful only when the
rollout depth is not too small.

\subsection{Generalization Across Instance Sizes}
\label{sec:exp:generalization}
We train at $10 \times 10$ and evaluate at $15 \times 10$ as a controlled
generalization probe. Gated selection remains substantially below Random-HH in
RPD and close to the best fixed rule, preserving the ordering observed in the
main test. This experiment is a diagnostic check rather than a full transfer
benchmark; it verifies that the selector is not tuned only to one instance
size.

\section{Discussion}\label{sec:discussion}

The fixed-rule results clarify the role of the gate. \textsc{FIFO} is strong
on these synthetic distributions, the experiments report it directly, and the
gate uses the best training-set rule as its default. The gain of gating is
therefore evaluated through its ability to avoid both random rule choice and
over-aggressive learned switching, and \emph{Regret-Gated} is consistently
stronger than the learned alternatives.

The comparison with deep-RL constructors such as L2D~\cite{zhang2020l2d} and
ScheduleNet~\cite{park2021schedulenet}, and with LLM-driven program search such
as FunSearch~\cite{romera-paredes2024funsearch} and ReEvo~\cite{ye2024reevo},
should be read in this context. Those systems target a different point in the
design space, while our pipeline keeps the model small and the rule pool fixed
so that label cost and switching reliability can be isolated cleanly.

From a deployment perspective, the strongest fixed rule in our synthetic JSSP
data is known only after running the comparison. In a new domain, a
practitioner may not know in advance whether \textsc{FIFO}, \textsc{MOPNR}, or
another rule is the right fallback. The proposed selector does not remove the
need for good rules, but it reduces the risk of using a learned selector that
switches too often. This objective is different from pure solver competition
and is relevant when interpretability and low deployment cost matter.

The study should also be read within a bounded scope. It is a controlled test
of rollout-calibrated rule selection rather than a claim about all scheduling
settings. The instances remain in small and medium regimes; scaling the same
labeling procedure to industrial JSSP, including $50 \times 20$ instances and
larger, may expose different cost--quality trade-offs. The rule pool is also
kept fixed to seven dispatching rules, which isolates selector behavior but
does not cover rules produced by LLMs or rule-induction systems. Because the
benchmark assumes static jobs and deterministic processing times, online
arrivals, machine breakdowns, or uncertain durations would change the state
distribution and would require rollout labels built for those dynamics.

\section{Conclusion}\label{sec:conclusion}
This paper studied how rollout supervision can be used in learning-assisted
JSSP without making a selector too eager to abandon a strong dispatching rule.
The proposed rollout-calibrated hyper-heuristic converts rollout outcomes into
per-state regret labels and uses a KNN uncertainty gate to switch only when the
predicted gain is sufficiently reliable. On the tested synthetic instances,
this conservative selector achieves the lowest mean RPD among learned methods,
reduces Random-HH RPD by more than an order of magnitude, and remains close to
the best fixed rule. The rollout-budget sweep further shows how label cost
translates into test quality, which is useful when interpretability, CPU-only
deployment, and simulation budget matter.

\bibliographystyle{named}
\bibliography{refs}

\end{document}